\documentclass{article}
\usepackage{microtype}
\usepackage{graphicx}
\usepackage{subfigure}
\usepackage{booktabs} % for professional tables

\usepackage{hyperref}

\usepackage[accepted]{icml2019}

\icmltitlerunning{Hyperbolic Disk Embeddings for Directed Acyclic Graphs}

\usepackage{amsmath}
\usepackage{amssymb}
\usepackage{bm}

\expandafter\def\csname ver@subfig.sty\endcsname{}
\usepackage{svg}

\newtheorem{theo}{Theorem}
\newtheorem{prop}{Proposition}

\newcommand{\x}{{\bm{x}}}
\newcommand{\y}{{\bm{y}}}

\newcommand{\Real}{{\mathbb{R}}}

\begin{document}

\twocolumn[
    \icmltitle{Hyperbolic Disk Embeddings for Directed Acyclic Graphs}
    % \icmltitle{Submission and Formatting Instructions for \\
    %           International Conference on Machine Learning (ICML 2019)}
    
    
    
    \icmlsetsymbol{equal}{*}
    
    \begin{icmlauthorlist}
    \icmlauthor{Ryota Suzuki}{lapras}
    \icmlauthor{Ryusuke Takahama}{lapras}
    \icmlauthor{Shun Onoda}{lapras}
    % \icmlauthor{Aeiau Zzzz}{equal,to}
    % \icmlauthor{Bauiu C.~Yyyy}{equal,to,goo}
    % \icmlauthor{Cieua Vvvvv}{goo}
    \end{icmlauthorlist}
    
    \icmlaffiliation{lapras}{LAPRAS Inc., Tokyo, Japan}
    % \icmlaffiliation{to}{Department of Computation, University of Torontoland, Torontoland, Canada}
    % \icmlaffiliation{goo}{Googol ShallowMind, New London, Michigan, USA}
    % \icmlaffiliation{ed}{School of Computation, University of Edenborrow, Edenborrow, United Kingdom}
    
    \icmlcorrespondingauthor{Ryota Suzuki}{suzuki@lapras.com}
    % \icmlcorrespondingauthor{Cieua Vvvvv}{c.vvvvv@googol.com}
    % \icmlcorrespondingauthor{Eee Pppp}{ep@eden.co.uk}
    
    \icmlkeywords{Embedding, Hyperbolic Spaces}
    
    \vskip 0.3in
]



\printAffiliationsAndNotice{}  % leave blank if no need to mention equal contribution

\begin{abstract}
Obtaining continuous representations of structural data such as directed acyclic graphs (DAGs)
has gained attention in machine learning and artificial intelligence.
However, embedding complex DAGs in which both ancestors and descendants
of nodes are exponentially increasing is difficult.
Tackling in this problem, we develop {\it Disk Embeddings},
which is a framework for embedding DAGs into quasi-metric spaces.
Existing state-of-the-art methods, Order Embeddings and Hyperbolic Entailment Cones,
are instances of Disk Embedding in Euclidean space and spheres respectively.
Furthermore, we propose a novel method {\it Hyperbolic Disk Embeddings} to handle exponential growth of relations.
The results of our experiments show that our Disk Embedding models
outperform existing methods especially in complex DAGs other than trees.

\end{abstract}

\section{Introduction}
\label{sec:introduction}

% 記号オブジェクトの適切な特徴表現を得ることは、今や機械学習や人工知能における中心的な関心事です。
Methods for obtaining appropriate feature representations of a symbolic objects are currently a major concern in machine learning and artificial intelligence.
Studies exploiting embedding for highly diverse domains or tasks have recently been reported for
graphs \cite{Grover2016NSF, Goyal2018GET, Cai2018ACS},
knowledge bases \cite{Nickel2011ATM, Bordes2013TEM, Wang2017KGE},
and social networks \cite{Hoff2002LSA, Cui2018ASN}.

% central concern は強すぎな気が．”is of great interest” あたりが平和．

% こいつらは略
% \item text disambiguation \cite{Ganea2017DJE}
% \item word hypernymy  \cite{Shwartz2016IHD}
% \item textual entailment \cite{Rocktaschel2015REN}
% \item taxonomy induction \cite{Fu2014LSH}
In particular,
studies aiming to embed linguistic instances into geometric spaces
have led to substantial innovations in natural language processing (NLP)
\cite{Mikolov2013DRW, Pennington2014GGV, Kiros2015STV, Vilnis2014WRG}.
% 予め学習された単語埋め込みを新しいモデルの学習における初期値として用いることで、さまざまなタスクでのパフォーマンスが向上することが確認されました。現在では、単語埋め込みはNLPタスクに取り組む際に必要不可欠な要素になっています。
Currently, word embedding methods are indispensable for NLP tasks, because it has been shown that the use of pre-learned word embeddings for model initialization leads to improvements in various of tasks \cite{Kim2014CNN}.
In these methods, symbolic objects are embedded in low-dimensional Euclidean vector spaces such that the symmetric distances between semantically similar concepts are small.

In this paper, we focus on modeling asymmetric transitive relations and directed acyclic graphs (DAGs).
% 最もよく調査されている asymmetric transitive relations は hierarchies です。Hierarchies は木構造を用いて表現することができます。木構造は単一の root node を持ち、子供の数が root からの距離に応じて指数的に増加する特性を持ちます。Asymmetric relations のうち、このような性質をもたないものは木構造を用いて表現することができませんが、DAGsを用いれば表現することができます。
Hierarchies are well-studied asymmetric transitive relations that can be expressed using a tree structure.
The tree structure has a single root node, and the number of children increases exponentially according to the distance from the root.
Asymmetric relations that do not exhibit such properties cannot be expressed as a tree structure, but they can be represented as DAGs,
which are used extensively to model dependencies between objects and information flow.
% ↓必要なら頑張って引用して尺を稼ぐこともできる。
% For example,  Biology で \cite{Mayr1968}, Social Science で \cite{Dodds2003}, Comparative linguists で \cite{Campbell2013}, Ontologies で \cite{Antoniou2004} などがある。
% DAGsはオブジェクト間の依存関係や情報の流れをモデル化するために広く使われます。
% 例えば、genealogyでは、家系図は親子関係を edge とし family member を node としたDAGとして解釈することができます。同様にして、 distributed revision system (例：Git）の commit objects もDAGを形成します。
For example, in genealogy, a family tree can be interpreted as a DAG, where an edge represents a parent child relationship and a node denotes a family member~\cite{Kirkpatrick2011HVG}.
Similarly, the commit objects of a distributed revision system (e.g., Git) also form a DAG\footnote{
\hyperlink{https://git-scm.com/docs/user-manual}
{https://git-scm.com/docs/user-manual}
}.
% citation network では、 citation graphs はある文書をノードとし文書間の引用関係をエッジとしたDAGであるとみなすことができます。
In citation networks, citation graphs can be regarded as DAGs, with a document as a node and the citation relationship between documents as an edge \cite{Price1965NSP}.
% 因果関係においては、DAGは疫学的研究における交絡の制御や、因果関係の形式的な理解を研究するための手段として用いられています。
In causality, DAGs have been used to control confounding in epidemiological studies \cite{Robins1987AGA, Merchant2002DAG} and as a means to study the formal understanding of causation \cite{Spirtes2000CPS, Pearl2003CMR, Dawid2010BD}.

Recently, a few methods for embedding asymmetric transitive relations have been proposed. 
% Nickel and Kiera は symbolic objects を embedding する空間を
% euclidean spaces ではなく hyperbolic spaces にするという研究の方向性を切り拓いた。
% これは hyperbolic spaces が any weighted tree を almost preserving their metric しながら埋め込むことができるという性質を踏まえてのことである。
\citeauthor{Nickel2017PEL} reported pioneering research on embedding symbolic
objects in hyperbolic spaces rather than Euclidean spaces and proposed 
{\it Poincar\'e Embedding} \cite{Nickel2017PEL, Nickel2018LCH}.
This approach is based on the fact that hyperbolic spaces can embed
any weighted tree while primarily preserving their metric \cite{Gromov1987HG, Bowditch2005ACO, Sarkar2011LDD}.
% Vendrov et al. は partial order を Euclidean spaces に埋め込まんとする手法である Order Embeddings を開発しました。
\citeauthor{Vendrov2016OEI} developed {\it Order Embedding},
which attempts to embed the partially ordered structure of a
hierarchy in Euclidean spaces \cite{Vendrov2016OEI}.
% この手法は、symbolic object の relation を partial ordering の一種であるユークリッド空間のreversed product order に埋め込んだ。
This method embeds relations among symbolic objects in reserved product orders in Euclidean spaces, which is a type of partial ordering.
% これは、ユークリッド空間におけるorthant たちのinclusive relation として表される。
Objects are represented as inclusive relations of orthants in Euclidean spaces.
% さらに、 Poincare Embeddings と Order Embeddings を継承して、 
%Ganea は {\it geodesically convex entailment cones} を提案した。
%それは、 symbolic objects を  convex cones として hyperbolic spaces に埋め込む。
Furthermore, inheriting Poincar\'e Embeddings and Order Embeddings,
\citeauthor{Ganea2017DJE} proposed {\it Hyperbolic Entailment Cones},
which embed symbolic objects as convex cones in hyperbolic spaces \cite{Ganea2018HEC}.
% 彼らは Poincare ball モデルの原点を中心としたPolar coordinate を用いることで、原点から遠ざかる(=階層を下る)につれてchild nodes が指数的に増えるようなモデルを立てている。
The authors used polar coordinates of the Poincar\'e ball 
and designed  their cones such that the number of descendants
increases exponentially as one moves going away from the origin.
%This method models the property that number of child nodes exponentially increase as they away from the origin by using Polar coordinate centered on the origin of Poincare ball model.
% このため、彼らは一般のDAGを扱うことができると述べているものの、ancestor とdescendant の両方が指数的に増えるようなグラフを埋め込むのには適していない。
Although the researchers reported that their approach can handle any DAG,
their method is not suitable for embedding complex graphs
in which the number of both ancestors and descendants grows exponentially,
%although they say that Hyperbolic Entailment Cones can handle general DAGs, it is not suitable for embedding graphs that exponentially increase both ancestors and descendants.
%実験によって有効性を示したと供述しているが、そのデータセットも単一の root node が存在するようなものになっていて、一般のDAGに関して良い埋め込みが得られているとは言い難い。
and their experiments were conducted using only hierarchical graphs.
Recently, \citeauthor{Dong2019ECT} suggested a concept of embedding
hierarchy to $n$-disks (balls) in the Euclidean space~\cite{Dong2019ECT}.
% 彼らは明らかに円によって埋め込むことができる
% ことがすでに知られている木構造のみに対応した
% bottom up construction のアルゴリズムを提案したにすぎず、
% 一般のDAGに適用することはできない。
However, their method cannot be applied to generic DAGs,
as it is a bottom up algorithm that can only be applied
to trees, which is previously known to be expressed by
disks in a Euclidean plane \cite{Stapleton2011DED}.

In this paper, we propose {\it Disk Embedding},
a general framework for embedding DAGs in metric spaces, or more generally, quasi-metric spaces.
% このフレームワークは、様々な existing state-of-the-art の一般化となっており。例えば、Order Embeddings \cite{Vendrov2016OEI} and Hyperbolic Entailment Cones \cite{Ganea2018HEC} は、それぞれユークリッド空間、球面におけるDiskEmbedding と等価であることが示される。
Disk Embedding can be considered as a generalization of the aforementioned state-of-the-art methods:
Order Embedding \cite{Vendrov2016OEI} and Hyperbolic Entailment Cones \cite{Ganea2018HEC}
are equivalent to Disk Embedding in Euclidean spaces and spheres, respectively.
% さらに、我々は、このフレームワークをHyperbolic Geometry に拡張することで、hyperbolic DE を提案する。この手法は、Hyperbolic space の並進対称性を保ちながら、(遠くに行けばたくさんうめこめる: 言葉を選ぶ)性質を生かすため、先祖、子孫ともに指数的に増えるようなデータに対処することもできるモデルとなっております。
Moreover, extending this framework to a hyperbolic geometry, we propose {\it Hyperbolic Disk Embedding}.
Because this method maintains the translational symmetry of hyperbolic spaces and
uses exponential growth as a function of the radius,
it can process data with exponential increase in both ancestors and descendants.
% 我々は、一般のDisk Embedding に対するRiemannian SGD による学習の一般理論を構築しただけでなく、よく使われるgeometries (Euc, Sphere, Hyperbolic) にかんして、RSGD のclosed form expression を求めました。
Furthermore, we construct a learning theory for general Disk Embedding using the Riemannian stochastic gradient descent (RSGD)
and derive closed-form expressions of the RSGD for frequently used geometries,
including Euclidean, spherical, and hyperbolic geometries.

Experimentally, we demonstrate that our Disk Embedding models
outperform all of the baseline methods, especially for DAGs other than trees.
We used three methods to investigate the efficiency of our approach: Poincar\'e Embeddings \cite{Nickel2017PEL},
Order Embeddings \cite{Vendrov2016OEI} and Hyperbolic Entailment Cones \cite{Ganea2017DJE}.

Our contributions are as follows:
\begin{itemize}
    \item % Disk Embedding というmetric space にDAG を埋め込む際のgeneral なフレームワークを用意し、学習方法を体系化させた。
    To embed DAGs in metric spaces, we propose a general framework, Disk Embedding, and systemize its learning method.
    \item % Existing methods がDisk Embedding の一種であることを証明した。
    We prove that the existing state-of-the-art methods can be interpreted as special cases of Disk Embedding.
    \item % Disk Embedding を拡張してHyperbolic Disk Embedding を提案した。
    We propose Hyperbolic Disk Embedding by extending our Disk Embedding to graphs with bi-directional exponential growth.
    \item % Experiment で、Hyperbolic Disk Embedding が無双(予定) した。
    Through experiments, we confirm that Disk Embedding models outperforms the existing methods, particularly for general DAGs.
\end{itemize}

% The reminder of this paper is organized as follows.
% \dots % In Section 2, we discuss related work regarding hyperbolic and ordered embeddings. In Section 2, we introduce our model and algorithm to compute the embeddings. In Section 4 we evaluate the efficiency of our approach on large taxonomies. Furthermore, we evaluate the ability of our model to discover meaningful hierarchies on real-world datasets.

% ======================================================
% Hyperbolic spaces の引用シリーズ
% \cite{Billera2001GSP}
% \cite{Bronstein2017GDL}
% ======================================================

% ======================================================
% 木の良さを言う引用シリーズ
% \cite{Hamann2018TLH}
% \cite{Lamping1995AFT}
% \cite{Krioukov2009GFS, Cvetkovski2009HER}

% \cite{Shavitt2008HEI, Krioukov2010HGC, Blasius2016EES}
% ======================================================

% 最適化について
% \cite{Bonnabel2013SGD}

% \cite{Sala2018RPH}
% \cite{Cannon1997HG}

\section{Mathematical preliminaries}
\label{sec:preliminaries}

\subsection{Partially ordered sets}

As discussed by \cite{Nickel2018LCH},
a concept hierarchy forms a partially ordered set (poset),
which~is a~set~$X$ equipped with reflexive, transitive, and antisymmetric
binary relations $\preceq$.
We extend this idea for application to a general DAG.
Considering the reachability from one node to another in a DAG,
we obtain a partial ordering.

Partial ordering is essentially 
equivalent to the inclusive relation between certain subsets
called the {\it lower cone} (or {\it lower closure}),
$C_{\preceq}(x) = \left\{ y \in X \middle| y \preceq x \right\}$.
\begin{prop}
\label{prop:partial-order}
Let $(X, \preceq)$ be a poset. Then, $x \preceq y$ holds if and only if
$C_\preceq(x) \subseteq C_\preceq(y)$.
\end{prop}
%If $X$ is continuous space,
%lower cones will be a continuous subset with certain volume.
Embedding DAGs in continuous space can be interpreted as
mapping DAG nodes to lower cones with a certain volume that contains its descendants.

{\bf Order isomorphism.}
A pair of poset $(X, \preceq), (X', \preceq')$ is {\it order isomorphic}
if there exists a bijection $f: X \to X'$ that preserves the ordering,
i.e., $x \preceq y \Leftrightarrow f(x) \preceq' f(y)$.
We further consider that two embedded expressions of a DAGs
are equivalent if they are order isomorphic.

\subsection{Metric and quasi-metric spaces}

A metric space is a set in which a metric function
$d: X \times X \to \Real$ satisfying the following four properties
is defined:
{\bf non-negativity}: $d(\bm x,\bm y) \geq 0$,
{\bf identity of indiscernibles}: $d(\bm x,\bm y) = 0 \, \Rightarrow \, \bm x=\bm y$,
{\bf subadditivity}: $d(\bm x,\bm y) + d(\bm y,\bm z) \leq d(\bm x,\bm z)$,
{\bf symmetry}: $d(\bm x,\bm y) = d(\bm y,\bm x)$
for all $\bm x, \bm y, \bm z \in X$.

For generalization, we consider a {\it quasi-metric}
as a natural extension of a metric
that satisfies the above properties except for the symmetry \cite{Goubault-Larrecq2013NHT, Goubault-Larrecq2013QMS}.
In a Euclidean space $\Real^n$,
we can obtain various quasi-metrics as follows:
\begin{prop}[Polyhedral quasi-metric]
\label{prop:convex-metric}
Let $W = \{\bm w_i \}$ be a finite set of vectors in $\Real^n$ such that
%$\mathrm{coni}(W) = \Real^n$,
%where
%$\mathrm{coni}(W) := \left\{ \sum_i a_i \bm{w}_i \middle| a_i \geq 0 \right\}$.
$\mathrm{coni}(W) :=
\left\{ \sum_i a_i \bm{w}_i \middle| a_i \geq 0 \right\} = \Real^n$.
Let
\begin{equation}
\label{eq:polyhedral-quasimetric}
d_W(\x, \y) := \max_{i} \left\{ \bm{w}_i^\top (\x - \y) \right\}.
\end{equation}
Then $d_W$ is a quasi-metric in $\Real^n$.
\end{prop}
The assumption $\mathrm{coni}(W) = \Real^n$
in Prop.~\ref{prop:convex-metric}
is equivalent to the condition that
convex polytope spanned by $W$
contains the origin in its interior.
The shape of disk $d_W(\bm p, \bm x) \leq r$ for fixed $\bm p$ and $r$ is
a polytope whose facets are perpendicular to the vectors of $W$.
For instance,
let $\{\bm e_k\}$ be a standard basis
and $W=\{\pm \bm{e}_k\}_{k=1}^n$;
then, $d_W$ becomes a uniform distance
$d_\infty (\x,\y) = \max_k |x_k - y_k|$,
whose disks form hypercubes.
%and if $W = \{(s_1, \cdots, s_n) | s_i = \pm 1\}$, then
%$d_W$ is equal to L1 distance $d_1(\x, \y) = \sum_i |x_i - y_i|$.
An example of $d_W$
whose disks are ($n\!-\!1$)-simplices
appears in Theorem~\ref{theo:equiv-ord-emb} in Sec.~\ref{subsec:order-emb}.
Except for it,
the word quasi-metric in this paper can be replaced by metric
since metric is quasi-metric.

\subsection{Formal disks}
\label{subsec:formal-disks}

\begin{figure}[t]
    % \vskip 0.2in
    \begin{center}
    \centerline{
       \includegraphics[width=\columnwidth]{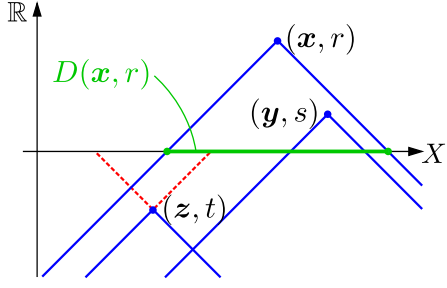}
    }
    \caption{
    Lower cones (solid blue lines) of generalized formal disks
    in $\mathbf B(X)$.
    $(\bm x,r) \sqsupseteq (\bm y,s), \, (\bm x,r) \sqsupseteq (\bm z,t)$
    holds and $(\bm z, t)$ has a negative radius $t<0$.
    }
    \label{fig:formal-disk}
    \end{center}
    \vskip -0.2in
\end{figure}

Let $(X,d)$ be a quasi-metric space and
consider a {\it closed disk}
$D(\x, r) = \left\{ \bm{p} \in X \middle| d(\bm{p}, \x) \leq r \right\}$,
which is a closed set in the sense of topology induced by $d$.
In considering a Euclidean space with ordinary Euclidean distances,
the inclusive relation of two closed disks is characterized as follows:
\begin{equation}
    D(\x, r) \supseteq D(\y, s) \Longleftrightarrow d(\x, \y) \leq r - s.
\end{equation}
Because this relation is a set inclusion relation,
it forms a poset, as discussed above.

In a general metric space,
we introduce {\it formal disks (balls)}\footnote{
This is generally called a formal ball,
but we call it a formal disk in this paper,
even in high-dimensional spaces for clarity.
},
which were first introduced by \citeauthor{Weihrauch1981EMS}
and studied as a computational model of metric space
\cite{Blanck1997DRM,Edalat1998CMM,Heckmann1999AMS}.
Formal disks are also naturally extended to
quasi-metric spaces~\cite{Goubault-Larrecq2013QMS}.
Let $(X, d)$ be a quasi-metric space,
where a formal disk is defined as a pair
$(\x, r) \in \bm{B^+}(X)=X\times \Real_+$
of {\it center} $\x$ and {\it radius} $r$.
The binary relation $\sqsupseteq$ of formal disks holds such that
\begin{equation} \label{eq:formal-disk}
    (\x,r) \sqsupseteq (\y,s)
    \Longleftrightarrow d(\x,\y) \leq r-s
\end{equation}
is a partial ordering \cite{Edalat1998CMM}.
%We can easily see that radius of formal disk is not necessarily nonnegative
%when defining partial order by \eqref{eq:formal-disk}.

When defining a partial order with \eqref{eq:formal-disk},
it is straightforward to determine that the radii of formal disks need not be positive.
We define $\bm{B}(X) = X \times \Real$
as a collection of {\it generalized} formal disks
\cite{Tsuiki2008LTS, Goubault-Larrecq2013NHT}
that are allowed to have {\it negative} radius.
Lower cones $C_\sqsubseteq(\x,r)$ of formal disks
are shown in Figure~\ref{fig:formal-disk}.
The cross section $\{(\bm x', 0) \in C_\sqsubseteq (\bm x, r) \}$
of the lower cone $C_\sqsubseteq(\x,r)$ at $r = 0$,
which is described as a solid green segment in Fig.~\ref{fig:formal-disk},
is a closed disk in X.
The negative radius of generalized formal disk can be regarded as
the radius of the cross section of the upper cone
(dashed red line in Fig.~\ref{fig:formal-disk}).

The following properties hold
for generalized formal disks.
\begin{description}
    \item[Translational symmetry.] For all $a \in \Real$,
    \vspace{-0.5em}
    \begin{equation} \label{eq:translational-symmetry}
    (\x,r) \sqsupseteq (\y,s) \Leftrightarrow (\x, r+a) \sqsupseteq (\y,s+a).
    \vspace{-0.5em}
    \end{equation}
    \item[Reversibility.] If $d$ is symmetric (i.e., $d$ is metric),
    \vspace{-0.5em}
    \begin{equation} \label{eq:reversibility}
    (\x,r) \sqsupseteq (\y,s) \Leftrightarrow (\x, -r) \sqsubseteq (\y,-s).
    \vspace{-0.5em}
    \end{equation}
\end{description}
These properties reflect the reversibility of the graph
and symmetry between generations,
which are important when embedding DAGs in posets of formal disks.
%(See Sec. \ref{sec:disk_embedding}).

\subsection{Riemannian manifold}

A Riemannian manifold is a manifold $\mathcal M$ with a collection of
inner products $g_{\bm{x}}: T_{\bm{x}} \mathcal M \times T_{\bm{x}} \mathcal M \to \Real$,
called a Riemann metric.
%in its tangent spaces $T_x \mathcal M$ for every points $x \in \mathcal M$.
Let $\gamma: [0,1] \to \mathcal M$ be a smooth curve,
where a length of $\gamma$ is calculated by:
\begin{equation*}
    \ell(\gamma) =
    \int_0^1 \sqrt{g_{\gamma(t)}(\gamma'(t), \gamma'(t))} \, dt.
\end{equation*}
The infimum of the curve length
$d_\mathcal{M}(\bm x, \bm y) = \inf_{\gamma} \ell(\gamma)$
from $\bm x$ to $\bm y$,
%of curves $\gamma$ such that $\gamma(0)=x, \gamma(1)=y$,
which becomes metric (and, consequently, quasi-metric) at $\mathcal M$.

{\bf Geodesic}:
A geodesic is a Riemannian analog of straight lines in Euclidean spaces
and is defined as a locally length-minimizing curve.
If Riemannian manifold is complete,
for every two points $\bm x$ and $\bm y$,
there exists a geodesic that connects $\bm x$ and $\bm y$ with a minimal length.

{\bf Exponential map}:
An exponential map $\exp_{\bm x}: T_{\bm x} \mathcal M \to \mathcal M$
is defined as $\exp_{\bm x}(\mathbf v) = \gamma(1)$,
where $\gamma$ is a unique geodesic satisfying
$\gamma(0) = x$ with an initial tangent vector $\gamma'(0) = \mathbf v$.
The map can also be interpreted as the destination reached after a unit time
when traveling along the geodesic from $x$ at an initial velocity of $\mathbf v$.
Therefore,
$
d(\bm{x}, \exp_{\bm x}(\mathbf v)) = \sqrt{g_{\bm x}(\mathbf v,\mathbf v)}
$
holds in a sufficiently small neighborhood of $\mathbf{v}=0$.

\subsection{Hyperbolic geometry}

A hyperbolic geometry is a uniquely characterized
complete and simply connected geometry
with constant negative curvature.
Hyperbolic geometries with curvature of $-1$
are described by several models
that are identical up to isometry.
The frequently used models in machine learning
include the Poincar\'e ball model \cite{Nickel2017PEL}
and the Lorentz model \cite{Nickel2018LCH}.

\begin{description}
\item[Poincar\'e ball model]
%Poincar\'e ball model
$\mathbb D^n$
is a Riemannian manifold defined on the
interior of an $n$-dimensional unit ball,
where distances are calculated as
\begin{equation}
d_\mathbb{D}(\x,\y) = \mathrm{arcosh}\left(
1 + 2 \frac{
    \lVert \x -\y \rVert ^{2}
}{
    (1-\lVert \x \rVert ^{2})(1-\lVert \y\rVert ^{2})
} \right),
\end{equation}
and its geodesics are arcs that perpendicularly intersect
the boundary sphere.
\item[Lorentz model] $\mathcal L^n$ is defined on
$n$-dimensional hyperboloid
$\left\{(x_0, \cdots, x_n) \middle| - x_0^2 + \sum_{k=1}^n x_k^2 = -1 \right\}$
embedded in $(n\!+\!1)$-dimensional Euclidean space,
where distance is calculated as
\begin{equation}
d_\mathcal{L} (\x, \y) = \mathrm{arcosh}(-\langle \x,\y \rangle_\mathcal{L}),
\end{equation}
where $-\langle \x,\y \rangle_\mathcal{L} = - x_0 y_0 + \sum_{k=1}^n x_k y_k$.
%Due to the equivalence of the Lorentz model and Poincar\'e ball model,
%They are mapped to each other via a diffeomorphism $p: \mathcal L \to \mathbb D$,
%$p(x_0, x_1, \cdots, x_n) = \frac{x_1, \cdots, x_n}{1 + x_0}$.
%\citeauthor{Nickel2018LCH}
\item[Translational symmetry]
Hyperbolic geometry is symmetric with respect to translations,
i.e., geometric properties including distances do not change
even if all points are translated in the same time.
In the Poincar\'e ball model, a translation that maps the origin to
$\bm y$ is obtained as
\begin{equation}
  \label{eq:hyperbolic-translation}
  \x \mapsto
  \frac{
    (1-\left\|\bm y \right\|^{2}) \, \bm{x}
    +(1+2 \left\langle \bm{x}, \bm{y}\right\rangle +\left\|\bm {x} \right\|^{2}) \, \bm y
  }{
    1+2 \left \langle \bm{x} , \bm{y} \right \rangle +\left\|\bm y \right\|^{2}\left\|\bm {x} \right\|^{2}
  },
\end{equation}
where $\langle \cdot, \cdot \rangle$ is an ordinary Euclidean inner product.

%Hyperbolic space は、負の曲率をもった
%
%Poincar\'e ball model
%
%Lorentz model
%
%isometry
%
%    transform と rotation
%    rotation は接空間でのrotation.
%    transform はPoincare ball ではこれ(数式)。

\end{description}

\section{Disk Embedding models}
\label{sec:disk_embedding}

In this section, we introduce {\it Disk Embeddings}
as a general platform for embedding DAGs
in quasi-metric spaces.
We first define Disk Embeddings
and evaluate its representability.
Second, we introduce Hyperbolic Disk Embeddings
and discuss its qualitative nature.
Third, we derive loss functions
and an optimization method
%to establish an expression
for Disk Embeddings
and finally obtain closed-form
expressions for some commonly used geometries.

\subsection{Disk Embeddings and its representability}

Let $C=\{c_i\}_{i=1}^{N}$ be a set of entities
with partial ordering relations $\preceq$,
and let $(X,d)$ be a quasi-metric space.
Disk Embeddings are defined as a map
$c_i \mapsto (\bm x_i, r_i) \in \mathbf B(X)$
such that $c_i \preceq c_j$ if and only if
$(\bm x_i, r_i) \sqsubseteq (\bm x_j, r_j)$.

\begin{figure}[t]
    \begin{center}
    \centerline{
        \includegraphics[width=\columnwidth]{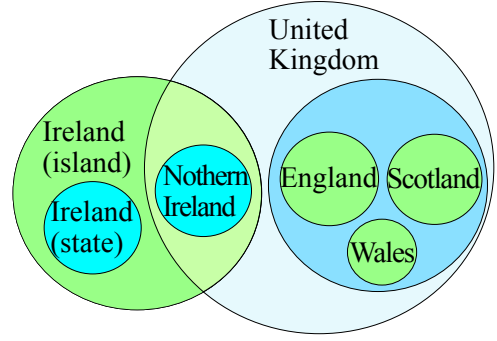}
    }
    \caption{
    Disk Embeddings in 2D Euclidean space with non-negative radii
    are equivalent to 2D Euler diagrams of circles.
    \label{fig:euler-diagram}
    }
    \end{center}
    \vspace{-0.5in}
\end{figure}

%例えば、$(X, d)$ を通常の
%ユークリッド平面 with ユークリッド距離とし、
%さらに、半径が正の場合に限ると、closed ball の包含関係で表される。
%これは、Euler diagram with circles で表していることと同じである
%ただし、inclusion だけが問題で、intersection はいらない。
%\cite{Stapleton2011DED}
%は、2D のEuler Diagram with circles のdrawability 
%(given な包含関係に対して)を研究して、
%certain class of diagrams named {\it pierced diagram }
%がDrawable なことを示した。
We can use various type of quasi-metric spaces $(X, d)$,
where $X$ can represent spherical or hyperbolic spaces,
and $d$ can be a unique quasi-metric.
Assuming that $X$ is an ordinary 2D Euclidean plane $\Real^2$
and that formal disks have positive radii,
Disk Embeddings become an inclusive relation of closed disks,
which is equivalent to an Euler diagram with circles,
except that only inclusions are meaningful and intersections do not make sense
(Figure~\ref{fig:euler-diagram}).
\citeauthor{Stapleton2011DED} studied the drawability of 2D Euler diagrams and demonstrated that
a certain class of diagrams, termed {\it pierced diagrams}, can be drawn with circles \cite{Stapleton2011DED}.
This result suggests that our Disk Embeddings will be effective
for certain classes of problems,
even for only three dimensions (two for centers and one for radii).
As \citeauthor{Dong2019ECT} mentioned,
the tree structure can be obviously embedded in $n$-disks because it is the simplest {\it pierced} diagram, in which none of circles intersect each other~\cite{Dong2019ECT}.
%3次元ユークリッド空間でのオイラー図は
%\cite{Rodgers2012I3V}によって導入されている。
%intersection を考慮する場合、2次元の円のみdrawable Euler diagram が
%3D でもdrawable であることはまだ証明されていないconjecture である。
%なぜなら、2次元を3次元にobvious な拡張は
%(中心を$(x_1, x_2) \mapsto (x_1, x_2, 0)$するような)
%intersection 関係を保たないから。
%pairwise inclusion のみを考慮するDisk Embeddingの場合、
%obvious extension to high dimension はpairwise inclusion
%を保つため、高次元の方が必ず高い表現力を持つ。
In higher Euclidean spaces,
Disk Embeddings have a greater representability than 2D
because all graphs that are representable in 2D can also be embedded 
in higher dimensions via a natural injection
$\iota:~\Real^2 \to \Real^n, $ s.t.
$\iota(x_1, x_2) = (x_1,x_2,0,\cdots,0)$.

% 関連するアイディアとしては、
% ミンコスキーtime space にDAGを埋め込む方法がCough2017 によって紹介された。
% ここでは、関連するノードが時間的関係にある。
% Disk embedding における lower (higher) cone は、
% みんこスキーtimespace における"light cone" に対応するものと考えられる。
As a related work,
embedding DAGs in Minkowski spacetime is introduced by
\cite{Clough2017EGL}, 
in which the related nodes are timelike separated.
The lower (upper) cones in Disk Embeddings can
be interpreted to the {\it light cones} of Minkowski spacetime.

Since we focus on transitive relation of DAGs,
non-transitive details of DAGs are not reserved
in any models based on order isomorphism
including Disk Embeddings, Hyperbolic Cones \cite{Ganea2018HEC} and
Order Embeddings \cite{Vendrov2016OEI}.
For instance, a DAG with three node $x, y, z$
and edges $x \to y$ and $y \to z$
will have a same representation as a DAG
having an additional edge $x \to z$.

\subsection{Hyperbolic Disk Embeddings}

We now introduce Disk Embeddings for a hyperbolic geometry.
%われわれは、さらに、双曲空間での Disk Embedding を考える。
In a hyperbolic geometry of two or more dimensions,
it is known that the area of the disk increases as an the
exponential function of the radius.
Thus, the lower cone section, shown in Figure~\ref{fig:formal-disk},
increases exponentially as the generation moves from
parent to child in the DAG.
In addition, it should be noted that both the inner
and outer sides of the lower cones becomes wider.
For instance, when $(\bm x, r) \sqsupseteq (\bm y, s)$,
the region $C_\sqsubseteq(\bm x, r) - C_\sqsubseteq(\bm y, s)$
%which is included in the lower cone of $(\bm u, r)$
%but is excluded from the lower cone of $(\bm v, s)$
also increases exponentially.
This property of the Hyperbolic Disk Embeddings
is suitable for embedding graphs in which
the number of descendants increases rapidly.

%2次元以上の双曲空間は、
%Diskの面積が半径の指数で増えることが知られている、
%かたや、ユークリッドではべきで増えるところを。
%これは、Fig.~\ref{fig:formal-disk}
%で説明したformal disk のlower cones の切り口が、
%親から子に世代を下るごとに指数関数的に大きくなっていく
%ことを意味している。
%付け加えて言えば、lower cones に含まれない領域もまた
%指数的に増加していくことに注意されたい。
%例えば、$(\bm u, r) \sqsupseteq (\bm v, s)$のとき、
%$(\bm u, r)$ の子孫ではあるが
%$(\bm v, s)$ の子孫ではない点の集合
%$C_\sqsubseteq (\bm u, r) -  C_\sqsubseteq (\bm v, s)$
%もまた、世代を下るごとに指数関数的に広がる。
%このことは、子孫が指数関数的に増えるような
%グラフを埋め込むのには適した性質である。

Considering the reversibility of
Disk Embeddings~\eqref{eq:reversibility},
the above property is established for not only the descendant
but also the ancestor direction.
In other words, not only lower cones but also upper cones
show an exponential extension.
In addition, considering translational symmetry
in hyperbolic spaces \eqref{eq:hyperbolic-translation} and
along radial axis \eqref{eq:translational-symmetry},
the same result holds for lower and upper cones starting from
any node in the graph.
Thus, in Hyperbolic Disk Embeddings,
complex threads that repeatedly intersect and separate from each other
in a complex DAG can be accurately expressed.

%上記で話した性質は、
%反転対称性\eqref{eq:reversibility}を考えると、
%子孫だけではなく、先祖方向にも同じことが言える。
%すなわち、lower cones だけでなく、upper cones も
%指数関数的に広がっているのである。
%また、Hyperbolic space 内と
%半径方向\eqref{eq:translational-symmetry}
%のtranslational symmetry を考えると、
%グラフ内の全てのノードを起点とした
%lower cones, upper cones に対して
%同様のことが成り立っている。
%これによって、
%Hyperbolic Disk Embedding では、
%複雑なDAGの中で多数平行して存在しいる
%親から子への複雑な系列
%(それらは、互いに交わったり離れたりを繰り返している)
%を正しく表現することができる。

\subsection{Loss functions}

We define a {\it protrusion parameter} between two formal disks
$(\bm x_i, r_i), (\bm x_j, r_j) \in \mathbf{B}(X)$
as
\begin{equation}
\label{eq:protrution}
l_{ij} = l(\bm x_i,r_i; \bm x_j, r_j) = d(\bm x_i, \bm x_j) - r_i + r_j.
\end{equation}

%このとき、
Because
$(\bm x_i, r_i) \sqsupseteq (\bm x_j, r_j)$ implies $l_{ij} \leq 0$,
%なので、
an appropriate representation can be obtained by learning
such that $l_{ij}$ is small for positive examples and 
large for negative examples.
%正例となるようなペアは$l$が小さくなるように、
%負例なるようなペアは$l$が大きくなるように学習することで、
%正しい関係性を学習することができる。

\begin{figure}[t]
    \vskip 0.2in
    \begin{center}
    \centerline{
        \includegraphics[width=3.5in]{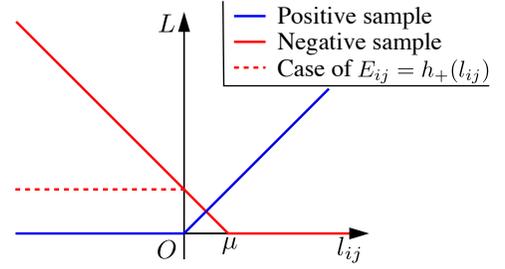}
    }
    \caption{
    Loss function.
    solid lines are the case of \eqref{eq:natural-energy}
    and dashed line is the case of $E_{ij} = h_+(l_{ij})$.
    \label{fig:relu}
    }
    \end{center}
    \vskip -0.2in
\end{figure}

Although various loss functions can be used,
we adopt the following margin loss in this study,
similar to that adopted by \cite{Vendrov2016OEI,Ganea2018HEC}:
%我々は、次のようなmargin loss
\begin{align}
\label{eq:loss-relu}
L = \sum_{(i, j) \in \mathcal T} h_+ (E_{ij})
+ \sum_{(i, j) \in \mathcal N} h_+ (\mu - E_{ij}),
\end{align}
where
$\mathcal{T}$ and $\mathcal{N}$ are sets of
positive and negative samples,
$\mu$ is a constant margin,
$h_+(x) = \max(0,x)$ and $E_{ij} = E(\bm x_i, r_i; \bm x_j, r_j)$
is an arbitrary energy function such that
$E_{ij}>0$ if and only if $l_{ij} > 0$.
The shape of \eqref{eq:loss-relu} is shown in Figure~\ref{fig:relu}.
Naturally, we can simply set
\begin{equation}
\label{eq:natural-energy}
E_{ij} = l_{ij},
\end{equation}
in which case, the loss function corresponds to
ones
%the loss functions
used by \cite{Vendrov2016OEI} and \cite{Ganea2018HEC},
except they used $h_+(l_{ij})$ instead,
causing that the gradient vanishes at $l_{ij} < 0$ in negative samples
(dashed line in Fig.~\ref{fig:relu}).
%(see Section~\ref{sec:equivalence}).

%and this is related to loss functions
%このロス関数は、あとで見るように、
%\cite{Vendrov2016OEI} や\cite{Ganea2018HEC} が採用した
%margin loss と密接に関係している。
%($\ell < 0$ for negative example で微分がゼロにならないことをのぞいて。)

\subsection{Riemannian optimization}

Given that $X$ is a general Riemannian manifold,
we must account for the Riemannian metric when
optimizing the loss function.
In this case, we utilize the RSGD method \cite{Bonnabel2013SGD},
%Riemannian stochastic gradient descent (RSGD) method
which is similar to the ordinary SGD except that
the Riemannian gradient is used instead of the gradient and
an exponential map is used for updates.
%$X$ が一般の多様体であるとき、$\mathbf{B}(X)$でのロス関数を最適化する場合、
%通常のgradient-based な手法は使えない。
%この場合、Riemannian stochastic gradient descent (RSGD) method
%\cite{Bonnabel2013SGD} を用いる。
%これは、gradient の代わりにRiemannian gradient を使うこと、
%update にexponential map を使うこと以外は同じ。

To execute the RSGD on $\mathbf{B}(X) = X \times \Real$,
we must determine the Riemannian metric in the product space
and calculate the corresponding Riemannian gradient.
To maintain translational symmetry
Eq.\eqref{eq:translational-symmetry} along the radial axis,
the Riemannian gradient on $\mathbf{B}(X)$ should have the following form
\begin{equation}
{\nabla}_{\mathbf{B}(X)} f(\bm x,r)
= \left(\lambda \nabla_{X;\bm{x}}
f(\bm x, r), \, \nu \frac{\partial}{\partial r} f(\bm x, r)\right),
\end{equation}
where $\nabla_{X;\bm{x}}$ is the Riemannian gradient on $X$
for variable $\bm x$
and $\lambda, \nu$ are positive constants.
%Furthermore, multiplying $\lambda$ and $\nu$ by a common positive
%corresponds to changing the learning ratio $\eta$; thus,
%we can set $\lambda=1$ without a loss of generality.
%RSGD を$\mathbf{B}(X) = X \times \Real$ で行うためには、
%この上のリーマン計量を求める必要がある。
%translational symmetry \eqref{eq:translational-symmetry} が成り立つためには、
%$\nabla_{\mathbf{B}(X)}
%= \left(\lambda \nabla_{\bm u}, \nu \frac{\partial}{\partial r}\right)$
%とすれば良い。
%where $\nabla_{\bm u}$ はRiemannian gradient operator on $\bm u \in X$ and
%$\lambda, \nu$はpositiveな定数。
%また、$\lambda, \nu$
%両方に正の定数を掛けるのは学習率を変更するのと同じだから、
%$\lambda = 1$ としてしまっても一般性を失わない。
Then the update formulae of RSGD for the parameters
%$\Theta_{i} = (\bm x_i, r_i)$ is given as follows:
$(\bm x_i, r_i)$ are given as follows:
\begin{align}
    \label{eq:update_x}
    \bm x_i^{(t+1)} &= \exp_{\bm x_i^{(t)}}\left[
        -\eta \lambda \nabla_{X; \bm x_i}  L^{(t)} \right], \\
    \label{eq:update_r}
    r_i^{(t+1)} &=  r_i^{(t)} - \eta \nu \frac{\partial L^{(t)}}{\partial r_i},
\end{align}
where $\eta$ is a learning ratio.

% 実際の形は、各geometry における metric 関数d(x,y) のgradient から、
%式(9),(10),(11)を通して簡単に計算される。
A specific form of \eqref{eq:update_x} is directly derived from
the Riemannian gradient of quasi-metric $d(\bm x, \bm y)$
and the exponential map for each geometry
via equations \eqref{eq:protrution} to \eqref{eq:natural-energy}.
%これは必ずしも必要なことではないが、
%もし
%quasi-metric $d$ of quasi-metric space $(X, d)$
%が$X$ のリーマン計量からinduce されるdistance $d_X$ と等しい場合、
%Although this is not always the case for general Disk Embeddings,
Especially,
%if the quasi-metric $d$ for formal disks is equal to the distance $d_X$ induced from
%the Riemannian metric in $X$,
if the distance function $d_X$ induced from the Riemannian metric in $X$ is used
as the quasi-metric $d$ for formal disks in \eqref{eq:protrution},
the Riemannian gradient of $d$ has a special form:
\begin{equation}
    \nabla_{X;\bm x} d(\bm x, \bm y) = \nabla_{X; \bm x} d(\bm y, \bm x) = - \gamma'(0)
\end{equation}
where $\gamma'(0)$ is the initial tangent vector of
the unit speed geodesic connecting from $\bm x$ to $\bm y$.
%ただし、$\gamma'(0), \gamma'(1)$ は、$\bm u$と$\bm v$とを結ぶ
%unit speed geodesic の始点と終点でのvelocity です。
%In this case, gradient of protrusion $l_{i,j}$ for $i, j$-th disk is calculated as follows,
%\begin{align}
%\nabla_{\mathbf{B}(X)}^{(i)} \, l_{ij} = \left(-\gamma'(0), \, - \nu\right), \,
%\nabla_{\mathbf{B}(X)}^{(j)} \, l_{ij} = \left(\gamma'(1), \,\nu\right).
%\end{align}
%%The RSGD algorithm is almost same as \cite{Nickel2018LCH}.
In addition, for frequently used geometries,
we here present closed form expressions as follows.

%Riemannian Optimization のアルゴリズム全体はAlgorithm~\ref{alg:rsgd}にまとめました。

\begin{description}
\item[Euclidean geometry.]
%{\bf Euclidean geometry:}
The RSGD in Euclidean spaces is equivalent to the standard SGD,
where the gradient of the distance is
$\nabla_{\bm x} d(\bm x, \bm y)
= \frac{\bm x - \bm y}{\| \bm x - \bm y\|}$
and the exponential map is
$\exp_{\bm x}(\mathbf v) = \bm x + \mathbf v$.

\item[Spherical geometry.]
%{\bf Spherical geometry:}
The RSGD on an $n$-sphere
$
\mathcal{S}^n
%:= \left\{\bm u \in \Real^{n+1} \,
%\middle| \, \|u\|_2=1 \right\}
$
with spherical distance $d_\mathcal{S}$
is conducted using the following formulae,
\begin{align}
\nonumber
&\nabla_{\mathcal S; \bm{x}} \, d_{\mathcal S}(\bm x, \bm y)
= \frac{ - \bm h}{\|\bm h\|}, \;\;
\bm h
= \bm y - \langle \bm x, \bm y \rangle \, \bm x, \\
\nonumber
&\exp_{\mathcal S; \bm{x}}(\mathbf v)
= \bm x \, \cos(\|\mathbf v\|)
    + \frac{\mathbf v}{\| \mathbf v \|} \, \sin(\|\mathbf v\|).
\end{align}

\item[Hyperbolic geometry.]
%{\bf Hyperbolic geometry:}
We use the Lorentz model \cite{Nickel2018LCH} for
Hyperbolic Disk Embeddings because
the RSGD in the Lorentz model involves considerably simpler formulae
and has a greater numerical stability
compared to the Poincar\'e ball model.
In the Lorentz model, the gradient of the distance and the exponential map are
computed as follows,
\begin{align}
\nonumber
&\nabla_{\mathcal{L};\bm{x}} \, d_{\mathcal L}(\bm x, \bm y)
= \frac{-\bm h}{\|\bm h\|_\mathcal{L}}, \;\;
\bm h
= \bm y + \langle \bm x, \bm y \rangle_\mathcal{L} \, \bm x, \\
\nonumber
&\exp_{\mathcal{L};\bm{x}}(\mathbf v)
= \bm x \, \cosh(\|\mathbf v\|_\mathcal{L})
    + \frac{\mathbf v}{\| \mathbf v \|_\mathcal{L}} \, \sinh(\|\mathbf v\|_\mathcal{L}),
\end{align}
where
$
\| \mathbf{v} \|_\mathcal{L} = \sqrt{\langle \mathbf v, \mathbf v \rangle_\mathcal{L}}.
$
%where
%$$\langle \bm u,\!\bm v \rangle_\mathcal{L}\!=\!-u_0 v_0\!+\!\sum_{i=1}^n u_i v_i, \;
%\|\bm u\|_\mathcal{L} = \sqrt{\langle \bm u,\!\bm u \rangle_\mathcal{L}}.$$

\end{description}

\section{Equivalence of Disk Embedding models}
\label{sec:equivalence}

\begin{figure}[t]
    \vskip 0.2in
    \begin{center}
    \centerline{
        \includegraphics[width=\columnwidth]{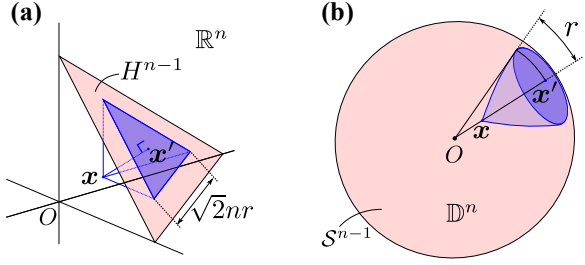}
    }
    \caption{
    Equivalence of Disk Embeddings and existing methods.
    {\bf(a)} Order Embeddings \cite{Vendrov2016OEI} and
    {\bf(b)} Hyperbolic Entailment Cones \cite{Ganea2018HEC}.
    \label{fig:embedding-equivalence}
    }
    \end{center}
    \vskip -0.2in
\end{figure}

In this section, we illustrate the relationship between
our Disk Embeddings and the current state-of-the-art methods.
We demonstrate that the embedding methods for DAGs in metric spaces
are equivalent to Disk Embeddings by projecting lower cones
into appropriate $n-1$ subspaces.

%state-of-the-art methods for embedding DAGs into metric space,
%State-of-the-art methods であるところの
%Order Embedding \cite{Vendrov2016OEI} と
%\cite{Ganea2018HEC} は、彼らの埋め込む空間における
%lower cones を、適切なn-1 次元subspace に射影することで、
%n-1次元のDisk Embeddings と等価であることを示す。

%以下では、
%ユークリッド空間に polyhedral distance
%\eqref{eq:polyhedral-quasimetric}
%を使ったDisk embedding がOrder Embedding \cite{Vendrov2016OEI}
%と同じで、
%球面におけるdisk embedding がHyperbolic Cones \cite{Ganea2018HEC}
%と同じことを示す。

%Disk embedding のLower cones は
%\ref{fig:formal-disk} に描き表した通りだが、
%\ref{subsec:formal-disks}で議論したように、
%Cone の包含関係とDisk の包含関係が、
%射影によって対応している。
%この考えを推し進めて、別の空間における
%order embedding のlower cones を
%適当なsubspace に射影することで、
%様々なモデルがDisk Embedding の枠組みで
%捉えることができると言うことを示す。

\subsection{Order Embeddings \cite{Vendrov2016OEI}}
\label{subsec:order-emb}

\citeauthor{Vendrov2016OEI}
proposed Order Embeddings, in which a
partial ordering relation is embedded in
reversed product order on $\Real_+^n$,
\begin{equation}
\label{eq:order-emb}
\x \succeq_\mathrm{ord} \y \; \Leftrightarrow \; x_k \leq y_k \; \text{for} \; k=1,\cdots, n.
\end{equation}
In Order Embeddings, the shape of the lower cone
$C_{\preceq}(\x) = [x_1, \infty)\times \cdots \times [x_n, \infty)$ is orthant.
%半順序構造の埋め込みを学習するという手法を提案した。
%彼らは、積順序(?)
%\begin{equation}
%\label{eq:order-emb}
%\x \succeq_\mathrm{ord} \y \; \Leftrightarrow \; \bigwedge_{i=1}^n x_i \leq y_i
%\end{equation}
%を使ったので、
%lower cones の形状はorthant になる。

As shown in Figure~\ref{fig:embedding-equivalence}({\bf a}),
we consider a projection onto a hyperplane
$
H^{n-1} = \left\{(x_1, \cdots, x_n) \middle| \frac{1}{n} \sum_{k=1}^{n} x_k=a \right\},
%\cong \Real^{n-1}
$
that is isometric to $\Real^{n-1}$.
The shape of the cross-section of the lower cone with $H^{n-1}$ is $(n\!-\!1)$-simplex%
\footnote{
    1-simplex is a line segment,
    2-simplex is a regular triangle,
    3-simplex is a regular tetrahedron,
    and so on.
}
%(Figure~\ref{fig:embedding-equivalence}({\bf a}))
Thus, we can consider the relation \eqref{eq:order-emb}
as an inclusive relation between corresponding simplices.
By using a polyhedral quasi-metric $d_W$ \eqref{eq:polyhedral-quasimetric},
we show that this is equivalent to Disk Embeddings in $H_n$ with
an additional constraint $r < a - d_W(\bm 0, \bm x)$, which forces
all entities to be descendants of the origin.
%を、$\Real^n$ のhyperplane とする。
%このとき、明らかに$H^{n-1} \cong \Real^{n-1}$(線型空間として同型)である。
%lower cones を$H^{n-1}$ で輪切りにしたcross section を考えてみると、
%その形はsimplex である(Fig.~\ref{fig:embedding-equivalence}{\bf a})。
%order embedding は、サイズの違うsimplex の包含関係で
%半順序を埋め込むものと考えることができる。
%simplex の包含関係というのは、Disk embedding の
%disk の包含関係と似てい。
%実際、quasi-metric として、polyhedral quasi-metric
%\eqref{eq:polyhedral-quasimetric}
%をつかうことで、order embedding を
%Disk embedding の一種と見ることができる。
\begin{theo}
\label{theo:equiv-ord-emb}
Order Embeddings \eqref{eq:order-emb} is
order isomorphic to
Euclidean Disk Embeddings with quasi-metric $d_W$
with $W\!=\! \{P \bm{e}_k \}_{k=1}^n$
via a smooth map $\phi_\mathrm{ord} \!:\! \Real_+^n \to
\mathbf B(H^{n-1})
%= H^{n-1} \times \Real
$,
\begin{equation}
\phi_\mathrm{ord}(\x) = (\bm x',r) = \left(P \x, \, a - \frac{1}{n} \sum_{k=1}^n x_k \right),
\end{equation}
where
%$\bm 1$ is a vector with all elements are one
$P = I - \frac{1}{n}\bm{1}\bm{1}^\top$ is a 
projection matrix onto $H^{n-1}$,
$\bm 1^\top = (1,\cdots,1)$.
\end{theo}

The energy function used for their loss function is
\begin{equation}
\label{eq:order-loss}
E^\mathrm{ord}(\x,\y) = \| h_+(\x-\y) \|^2,
\end{equation}
%which そのまますぐにprotrusion $l_{ij}$ \eqref{eq:protrution}
%で表すことはできないが、次のようなlower bound でよく近似することができる。
which cannot be directly expressed using $l_{ij}$ defined in \eqref{eq:protrution},
but can be well approximated by a lower bound.
\begin{theo}
\label{theo:loss-ord-emb}
Energy function \eqref{eq:order-loss} has a lower bound:
\begin{equation}
  \label{eq:order-loss-approx}
  E^\mathrm{ord}_{ij} \geq h_+(l_{ij})^2,
\end{equation}
and the equality holds iff
$\left| \left\{k \middle|
\left(\bm x_j - \bm x_i\right)_k > 0
\right\} \right| \leq 1$.
\end{theo}
In contrast to \eqref{eq:natural-energy},
\eqref{eq:order-loss-approx} has a quadratic form, which may cause
optimization being exponential decay
%$l_{ij} \to +0$
of $l_{ij}$
instead of crossing zero and
a vanishing gradient in $l_{ij} < 0$ even for negative samples,
making the optimization inefficient.

\subsection{Hyperbolic Entailment Cones \cite{Ganea2018HEC}}

\citeauthor{Ganea2018HEC} developed a method for
embedding cones extending in the radial direction of
the Poincar\'e ball model.
The embedding relation is expressed as follows
%原点を中心に動径方向にのびたcone をembedする。
%cone の包含関係で順序を表現している。
%彼らの順序関係は、次の式で表される。
\begin{equation}
\label{eq:hyp-cones}
    \x \succeq_\mathrm{hyp} \y \; \Leftrightarrow \; \psi(\x) \geq \Xi(\x, \y),
\end{equation}
where $\Xi(\x, \y)$ is the angle $\pi - \angle Oxy$ and
$\psi(\x) = \arcsin\left(K\frac{1-\|\x\|^2}{\|\x\|}\right)$
%is the angle between the axis and the generatrix of the cone.
is the generatrix angle of the cone.
%はconeの開き方の角度。

The authors focused on the polar coordinates of the Poincar\'e ball,
in which rotational symmetry around the origin is assumed
and the position in the hierarchical structure is mapped to the radial component.
%Thus, they treated the hyperbolic geometry as the
%product $\mathcal S^{n-1} \times R_+$,
%indicating that
Thus,
they implicitly assumed a non-trivial separation of
the hyperbolic space into $\mathcal S^{n-1} \times \Real_+$.
To illustrate this point, we consider projections of entailment cones onto the border of
a Poincar\'e ball $\partial \mathbb D \cong \mathcal S^{n-1}$.
As shown in Figure~\ref{fig:embedding-equivalence}({\bf b}),
the projections of the entailment cones form disks in an $(n\!-\!1)$-sphere,
and relation \eqref{eq:hyp-cones} is represented as the inclusion of corresponding disks.
%彼らの手法は原点中心の回転に着目しており、
%動径方向の成分はコーンの開き方にのみ対応しているため、
%実際には彼らの手法は球面へのdisk Embedding と等しい。
%このことを見るために、彼らのcones のPoincare` ball の縁への射影を考える。
%たとえば、3次元の場合、
%Figure~\ref{fig:embedding-equivalence}({\bf b}) のように、
%entailment cones のpoincare` ball の縁への射影は、
%球面上のdisk になっており、
%entailment cones の包含関係はdisk の包含関係と同値になっていることがわかる。
\begin{theo}
\label{theo:equiv-hyp-cones}
Hyperbolic Cones \eqref{eq:hyp-cones} are
order isomorphic to Disk Embeddings on $(n\!-\!1)$-sphere 
$\mathcal S^{n-1}$
via a smooth map: $\phi_\mathrm{hyp}: \mathbb D^n \to
\mathbf{B}(\mathcal S^{n-1})
%= \mathcal S^{n-1}\times \Real
$,
\begin{align}
\phi_\mathrm{hyp}(\bm x) &= (\bm x',r)
\nonumber \\
&=
\left(
    \frac{\x}{\|\x\|}, \;
\arcsin\!\left(\frac{1+\|\x\|^2}{2\|\x\|}\sin \theta_0 \right)-\theta_0
\right),
\label{eq:hyp-cones-map}
\end{align}
%\vskip -15pt
where $\theta_0 = \arctan{2K}$.
\end{theo}

%彼らのenergy function は、
Their energy function is given as 
\begin{equation}
\label{eq:loss-hyp-cones}
E^\mathrm{hyp}(\x, \y) = h_+\left(\Xi\left(\x,\y\right)
- \psi\left(\left\|\x\right\|\right)\right),
\end{equation}
and is also represented in the format of Disk Embeddings,
by using only distances between centers and the radii of formal disks.
%で表されるが、これは、Disk Embedding の描像では、次のように表される。
\begin{theo}
\label{theo:loss-hyp-cones}
Energy function \eqref{eq:loss-hyp-cones} has the following form:
\begin{equation}
\label{eq:loss-hyp-cones-equiv}
E^{\mathrm{hyp}}_{ij}
= h_+ \left(
\arcsin \!\left( q(d_{ij}, r_i, r_j) \cdot 2\sin\!\left(\frac{l_{ij}}{2}\right)\right)
\right),
\end{equation}
where
\begin{align}
q(d_{ij},r_i,r_j) =&
    \frac{
        \cos \left(\frac{r_i + r_j - d_{ij}}{2} + \theta_0 \right)
    }{
        \cos \theta_0 \, \sin \theta_0
    }
\\
\nonumber 
&\times \sqrt{
    \frac{
        \sin r_i \, \sin(r_i+2 \theta_0)
    }{
        s_i^2 + s_j^2 - 2 s_i s_j \cos d_{ij} - \sin^2 d_{ij}
    }
},
\end{align}
and $s_i = \frac{\sin(r_i+\theta_0)}{\sin \theta_0},
s_j = \frac{\sin(r_j+\theta_0)}{\sin \theta_0}$.
\end{theo}
As can be easily seen from \eqref{eq:loss-hyp-cones-equiv},
the energy
$E_{ij}^{\mathrm{hyp}}$ is linearly approximated by:
\begin{equation}
\label{eq:loss-hyp-cones-approx}
E_{ij}^{\mathrm{hyp}} = q(r_i-r_j, r_i, r_j) \cdot h_+(l_{ij})
+ \mathcal{O}\left(d_{ij}^2\right),
\end{equation}
%\begin{equation}
%q(r_i-r_j, r_i, r_j) =
%\frac{
%    \cos \left(r_j + \theta_0 \right) \sqrt{\sin r_i \sin(r_i + 2\theta_0)}
%}{
%    \cos^2 \theta_0 \cdot \sin(r_i - r_j)
%}
%\end{equation}
around $d_{ij} \approx r_i - r_j$ for fixed $r_i, r_j$.
%(Figure~\ref{fig:hyp-cones-loss}).
Equation~\eqref{eq:loss-hyp-cones-approx} is
similar to \eqref{eq:natural-energy}
except that the coefficient $q(r_i-r_j,r_i,r_j)$ and
gradient vanishing at $l_{ij}<0$ as observed for \eqref{eq:order-loss-approx}.

%Disk embedding の描像$(\bm{u}_i, r_i)$では、この目的関数は
%中心間の距離$d_{ij} = d_S(u_i, u_j)$ と半径$r_i$のみを用いて次のように表すことができる。
%\begin{strip}
%\end{strip}

%\begin{table*}[h]
%\begin{center}
%\begin{align}
%\Xi_{ij} - \psi_i =
%2 &\sin\left( \frac{r_i - r_j - d_{ij}}{2}\right) \cdot
%\frac{\cos \left(\frac{r_i + r_j - d_{ij}}{2} + \theta_0 \right) }{\cos \theta_0}
%\nonumber \\
%&\times \sqrt{\frac{
%    \sin r_i \sin(r_i+2 \theta_0) 
%}{
%    \sin^2(r_i+\theta_0) + \sin^2(r_j+\theta_0)
%    - 2 \sin(r_i+\theta_0) \sin(r_j+\theta_0)\cos d_{ij}
%    - \sin^2 d_{ij} \sin^2 \theta_0
%}}
%\end{align}
%\end{center}
%\end{table*}

%\begin{table*}[t]
%\caption{Disk and point embedding models}
%\label{sample-table}
%\vskip 0.15in
%\begin{center}
%\begin{small}
%%\begin{sc}
%\begin{tabular}{l|ccc}
%\toprule
%& Spherical Space & Euclidean Space & Hyperbolic Space \\
%\midrule
%Point Embedding & - & Word2Vec & 
%\begin{tabular}{c}
%    \cite{Nickel2017PEL} \\ \cite{Nickel2018LCH}
%\end{tabular} \\
%\midrule
%Disk Embedding & \cite{Ganea2018HEC} & \cite{Vendrov2016OEI} & Our Hyperbolic GDE \\
%\bottomrule
%\end{tabular}
%%\end{sc}
%\end{small}
%\end{center}
%\vskip -0.1in
%\end{table*}

\subsection{Vector Embedding models}

Ordinary embedding models based on similarity,
e.g, word2vec~\cite{Mikolov2013DRW} and
Poincar\'e Embeddings~\cite{Nickel2017PEL,Nickel2018LCH},
can be seen as a application of Disk Embeddings
in which radius information is neglected.

\citeauthor{Nickel2018LCH} argued that
when embedding  in hyperbolic spaces,
general concepts can be obtained closer to the origin
by learning with loss function distances between points.
However, because a hyperbolic space has translational symmetry,
there are no special point and any point can be the origin.
Thus,  simultaneously translating all of the points
does not change the loss function,
and which node is closer to the origin is determined only by the initial vectors.
%all the points can be closer to the origin,
%and initial vectors of learning determine it.
Furthermore, an approach in which distances from the origin are interpreted
as levels in a hierarchy is not suitable for 
complex DAGs in which both ancestors and descendants grow exponentially,
with no single root.

%\cite{Nickel2018LCH} は、
%Hyperbolic space において、
%2点間の距離のみを用いたロス関数で学習することで、
%原点に近いほど(経験的には)一般的な概念が得られる
%(なぜなら、一般的な概念は多くのノードとのconectivityを持つから)
%と主張している。
%しかしながら、
%hyperbolic space において
%平行移動はisometric なので、
%全ての点を同時に平行移動しても
%loss function の値は変化しない。
%したがって、全てのentity は
%その一般性やconnectivityによらず原点に近くなりうるし、
%それは、学習の初期状態の与え方によって決まってしまうものである。
%さらには、このような原点からの距離を
%階層構造におけるlevel と見るheuristics は、
%complex DAGs where both ancestors and descendant
%grows exponentially では適用できない。
%なぜなら、そこには、単一のルートのようなものはないからである。

\section{Experiments}
\label{sec:experiments}

In this section we evaluate  Disk Embedding models
for various metric spaces including Euclidean,
spherical and hyperbolic geometries.%using existing methods.

\begin{table}[t]
\vspace{-0.05in}
\caption{
Dataset statistics.
%The number of nodes, edges and
The average number of ancestors (descendants)
of leaf (root) nodes are shown.
}
\label{tab:dataset-stat}
% \vskip 0.15in
\begin{center}
\begin{small}
%\begin{sc}
\begin{tabular}{lccrr}
\toprule
  Dataset & Nodes & Edges & \shortstack{Ancestors} & \shortstack{Descendants} \\
\midrule
  WordNet & 82,115 & 743,086 & 9.2 & 82114.0  \\
  %HEP-TH & 27,739 & 351,389 & 4650.4 & 7474.9  \\
\bottomrule
\end{tabular}
%\end{sc}
\end{small}
\end{center}
\vspace{-0.2in}
\end{table}

\begin{table*}[t]
\caption{Test F1 results for various models. Hyperbolic Entailment Cones is proposed by~\cite{Ganea2018HEC}, Order Embeddings is proposed by~\cite{Vendrov2016OEI} and Poincar\'e Embeddings is proposed by~\cite{Nickel2017PEL}.}
\label{tab:results}
\vskip 0.15in
% \begin{sc}
\begin{center}
\begin{small}
\begin{tabular}{llcccccccccc}
\toprule
&& \multicolumn{4}{c}{Embedding Dimension $= 5$} && \multicolumn{4}{c}{Embedding Dimension $= 10$} & \\ \cline{3-6} \cline{8-11} \vspace{-1.5mm} \\
&& \multicolumn{9}{c}{Percentage of Transitive Closure (Non-basic) Edges in Training} & \\
& & 0\% & 10\% & 25\% & 50\% && 0\% & 10\% & 25\% & 50\% & \\ \midrule
\multicolumn{2}{l}{\bf WordNet nouns} \\
& Our Euclidean Disk Embeddings   & 35.6\% & 38.9\% & 42.5\% & 45.1\% && {\bf 45.6}\% & 54.0\% & 65.8\% & 72.0\% & \\
& Our Hyperbolic Disk Embeddings  & 32.9\% & 69.1\% & 81.3\% & 83.1\% && 36.5\% & 79.7\% & 90.5\% & {\bf 94.2}\% & \\
& Our Spherical Disk Embeddings   & {\bf 37.5}\% & {\bf 84.8}\% & {\bf 90.5}\% & {\bf 93.4}\% && 42.0\% & {\bf 86.4}\% & {\bf 91.5}\% & 93.9\% & \\
& Hyperbolic Entailment Cones      & 29.2\% & 80.0\% & 87.1\% & 92.8\% && 32.4\% & 84.9\% & 90.8\% & 93.8\% & \\
& Order Embeddings                & 34.4\% & 70.6\% & 75.9\% & 82.1\% && 43.0\% & 69.7\% & 79.4\% & 84.1\% & \\
& Poincar\'e Embeddings           & 28.1\% & 69.4\% & 78.3\% & 83.9\% && 29.0\% & 71.5\% & 82.1\% & 85.4\% & \\
\midrule
\multicolumn{2}{l}{\bf WordNet nouns reversed} \\
& Our Euclidean Disk Embeddings   & {\bf 35.4}\% & 38.7\% & 42.3\% & 44.6\% && {\bf 46.6}\% & 55.9\% & 67.3\% & 70.6\% & \\
& Our Hyperbolic Disk Embeddings  & 30.8\% & 49.0\% & 66.8\% & 78.5\% && 32.1\% & 53.7\% & 79.1\% & 88.2\% & \\
& Our Spherical Disk Embeddings   & 34.8\% & {\bf 59.0}\% & {\bf 76.8}\% & {\bf 84.9}\% && 38.0\% & 60.6\% & {\bf 83.1}\% & {\bf 90.1}\% & \\
& Hyperbolic Entailment Cones      & 17.3\% & 57.5\% & 71.8\% & 75.7\% && 20.5\% & {\bf 61.9}\% & 73.1\% & 75.8\% & \\
& Order Embeddings                & 32.9\% & 33.8\% & 34.8\% & 35.8\% && 34.7\% & 36.7\% & 38.8\% & 41.4\% & \\
& Poincar\'e Embeddings           & 26.0\% & 48.4\% & 48.8\% & 51.4\% && 27.4\% & 49.7\% & 50.9\% & 51.9\% & \\
\bottomrule
\end{tabular}
\end{small}
\end{center}
% \end{sc}
\vskip -0.1in
\end{table*}

\subsection{Datasets} \label{subsec:datasets}

For evaluation we use the 
{\it WordNet} \textregistered~\cite{Miller1995WN}\footnote{ 
\hyperlink{https://wordnet.princeton.edu/}
{https://wordnet.princeton.edu/}
}, a large lexical database that provides hypernymy relations.
In our experiments, we used a noun closure for
evaluating hierarchical data.
%
%%For evaluation we use the following network datasets.
%\begin{description}
%    \item[WordNet \textregistered] \cite{Miller1995WN}\footnote{ 
%    \hyperlink{https://wordnet.princeton.edu/}
%    {https://wordnet.princeton.edu/}
%    }
%      A large lexical database that provides hypernymy relations.
%      In our experiments, we used a noun closure for
%      evaluating hierarchical data.
%%    \item[arXiv HEP-TH]
%      %A high-energy physics theory (HEP-TH) citation network
%      %from the e-print arXiv,
%      %which was originally released as 
%      %part of the 2003 KDD Cup \cite{Gehrke2003OKC}\footnote{
%      %\hyperlink{http://www.cs.cornell.edu/projects/kddcup/}
%      %{http://www.cs.cornell.edu/projects/kddcup/}
%      %}.
%      %We used the data as a benchmark for embedding citation networks.
%    %\item[APS citation dataset]
%      %これは計算が間に合わないかも
%      %Available online with simple query\footnote{
%\end{description}
%
%各データの統計はTable~\ref{tab:dataset-stat}のようになっている。
%WordNet noun データは、ヒエラルキーの例で、
%子孫の数が非常に多いのに対し、
%先祖の数が非常に限られていることが特徴である。
%
The statistics for the dataset are shown in Table~\ref{tab:dataset-stat}.
The WordNet noun dataset is an example of
a tree-like hierarchical network,
characterized by a highly limited number of ancestors,
whereas the number of descendants is considerably large.
%
%HEP-TH は、非ヒエラルキーの例であって、
%子孫と先祖の両方がたくさん存在する。
%
%In contrast, HEP-TH dataset is an example of a more
%complicated network with many ancestors and descendants.
%We remove backdating edges from the HEP-TH dataset to maintain a DAG structure.
%%which results in 0.40 \% fewer edges.
%%transitive relation is 
%Further we sampled 1,000 nodes such that transitive relation is preserved,
%and a series of train data is built in the same way as
%\citeauthor{Ganea2018HEC} have done;
%we  extract {\it basic} edges which constitute transitive reduction of the graph,
%and generate train data by basic edges with 0\%, 10\%, 25\% or 50\% of the non-basic edges.
%
%我々は、さらに、各データセットの関係性を入れ替えたデータ
%も用意した。DAG のreverse はDAGであるため、
%これらのデータも非ヒエラルキーのDAG の例となっとる。
%
We also used data obtained by
inverting the relations in WordNet noun dataset.
Because the reverse of a DAG is a DAG,
%these data also are become examples
%of DAGs other than tree-like hierarchies.
this data is considered to be an example of 
non-tree structural DAG.

\subsection{Training and evaluation details}
%我々は
%これらのデータのエッジ
%をpositive な例$(i,j) \in \mathcal T$の例
%such that $c_i \succeq c_j$ としてがくしゅうした。
%これらのデータはpositive なデータしかないため、
%我々は、negative なデータをRSGD
%(Algorithm~\ref{alg:rsgd})
%の各iteration ごとにランダムにsampleすることで
%学習を行った。
We conducted learning by using pairs of nodes
connected by edges of these graph data
as positive examples $(i,j) \in \mathcal T$
such that $c_i \preceq c_j$.
Because these data only have positive pairs,
we randomly sampled negative pairs
for each iteration of RSGD.
%Our Spherical Disk Embeddings and Hyperbolic 
%Disk Embeddings are initialized using Poincar\'e Embeddings~\cite{Nickel2017PEL}
%as pre-training to make the same condition
%with Hyperbolic Entailment Cones~\cite{Ganea2018HEC}.
%
%我々は、学習された表現の評価方法として、
%ランダムに選ばれた2つのノード$(i,j)$
%がtransitive relation $c_i \succeq c_j$
%を満たすか否か
%(すなわち、DAG におけるノード$i$ から$j$への
%有向パスが存在するか否か)を当てる２値分類問題として
%F1 score を使って評価した。
%
For evaluating the learned expression,
we use an F1 score for a binary classification of
whether a randomly selected pair $(i, j)$
satisfies the transitive relation $c_i \succeq c_j$,
in other words, to assess whether there exists a
directed path from $i$ to $j$ in the DAG.
%
%テストデータは、ランダムに1000個のノードのペアを選んで、
%その中から重複とidentical なペア$(i,i)$ を除いて作った。
%
%The test data are generated by randomly sampling
%1,000 pairs of nodes while removing duplicates and
%identical pairs $(i,i)$.

\subsection{Baselines}
%我々は、対戦相手として、
%Poincar\'e Embedding \cite{Nickel2017PEL},
%Order Embedding \cite{Vendrov2016OEI} and
%Hyperbolic Cones \cite{Ganea2018HEC}
%を持ってきた。
%これらの手法は、\citeauthor{Ganea2018HEC}
%の実装を用いて実験を行った\footnote{
%\hyperlink{https://github.com/dalab/hyperbolic_cones}
%{https://github.com/dalab/hyperbolic\_cones}
%}.
%また、パラメータについても、
%彼らの実装と同じものを使った。
We also evaluated
Poincar\'e Embeddings~\cite{Nickel2017PEL},
Order Embeddings~\cite{Vendrov2016OEI} and
Hyperbolic Entailment Cones~\cite{Ganea2018HEC}
as baseline methods.
For these baseline methods,
we used the implementation reported by
\citeauthor{Ganea2018HEC}\footnote{
\hyperlink{https://github.com/dalab/hyperbolic_cones}
{https://github.com/dalab/hyperbolic\_cones}
}.
In addition, experimental conditions such as hyperparameters
are designed to be nearly similar to those of the experiments
conducted by \citeauthor{Ganea2018HEC}.
%
%我々は、
%\cite{Ganea2018HEC}がpretrain に\cite{Nickel2017PEL}
%を用いているので、これとのfair な比較のために、
%Spherical Disk Embedding で適用可能なときは
%我々もこれをpretrain に用いた。
%その場合、Nickel によるpretrain が終わった後、
%\eqref{eq:hyp-cones-map}
%によって球面上のFormal disk に変換してから、
%Disk Embedding のRSGD の学習をおこなった。
%
Considering~that Hyperbolic Cones \cite{Ganea2018HEC}
use Poincar\'e Embeddings~\cite{Nickel2017PEL} for pretraining,
we also apply this approach to Spherical Disk Embeddings
(which is equivalent to Hyperbolic Cones
as shown in Theorem~\ref{theo:equiv-hyp-cones})
to make a fair comparison.
%
%Poincar\'e Embedding \cite{Nickel2017PEL}
%はsimilarity ベースの対称な関係を学習する手法であるが、
%彼らが用いたheuristics であるスコアを用いることで
%asymmetric な関係の推定を行った。
%
Although Poincar\'e Embeddings~\cite{Nickel2017PEL}
is a method used for learning symmetric relations based on similarity,
it can also be used to estimate asymmetric relations
by using a heuristic score
%\begin{equation}
$S(\bm x, \bm y)
= (1 + \lambda(\|\bm x\| - \|\bm y\|)) \, d_{\mathbb{D}}(\bm x, \bm y)$,
%\end{equation}
%
%パラメータlambdaの値は、
%test data とは別にサンプルされた
%validation set の
%F1 measure を最大にするように、決定される。
%
where the parameter $\lambda$ is determined by
maximizing the F1 score for validation data
which are sampled separately from the test data.

\subsection{Results and discussion}
Table~\ref{tab:results} shows the F1 score on each dataset.
As shown in Section~\ref{sec:equivalence}, we proved equivalence between our Spherical Disk Embeddings and Hyperbolic Entailment Cones.
It is observed that our Spherical Disk Embeddings reaches almost the same result of Hyperbolic Entailment Cones with {\bf WordNet nouns}.
The slight improvement of our model can be explained the change of the loss function.
{\bf WordNet nouns reversed} is generated from WordNet by reversing directions of all edges,
nevertheless, it is an example of DAGs.
In this data, our Disk Embeddings models obviously outperformed other existing methods
because our methods maintain reversibility (Eq.~(\ref{eq:reversibility})) while existing methods implicitly assume hierarchical structure.
%Finally, according to {\bf HEP-TH} in Table~\ref{tab:results}, we can confirm that our methods overcome all existing methods and the efficiency on complex DAGs of out methods as well as {\bf WordNet nouns reversed}.

\section{Conclusion}
\label{sec:conclusion}

We introduced Disk Embeddings, which is a new framework for embedding DAGs in quasi-metric spaces to generalize the state-of-the-art methods.
Furthermore, extending this framework to a hyperbolic geometry, we propose Hyperbolic Disk Embedding.
%Experimentally, we demonstrate that Hyperbolic Disk Embeddings outperform all of the baseline methods, especially for general DAGs.
Experimentally we demonstrate that our methods outperform all of the
baseline methods, especially for DAGs other than tree.

For future work, large-scale experiments for complex DAGs
such such as citation networks is desired,
in which both ancestors and descendants increase rapidly,
and exponential nature of Hyperbolic Disk Embedding
will demonstrate its core.

For reproducibility, our source code for the experiments are
publicly available online\footnote{
\hyperlink{https://github.com/lapras-inc/disk_embedding}
{https://github.com/lapras-inc/disk-embedding}
}.
The datasets we used are also available online. See Section~\ref{subsec:datasets}.
%Our code for experiments is submitted in supplementary material.
%Reviews can reproduce our experimental results by executing the code in accordance with \verb|README.md|.
%It will be publicly available in the camera ready version.

%We conducted experiments using open datasets.
%As shown in Section~\ref{subsec:datasets}, anyone can 

\section*{Acknowledgements}
\label{sec:acknowledgements}
We would like to thank Tatsuya Shirakawa,
Katsuhiko Hayashi, and all the anonymous reviewers
for their insightful comments, advice and suggestions.

% We strongly encourage the publication of software and data with the
% camera-ready version of the paper whenever appropriate. This can be
% done by including a URL in the camera-ready copy. However, do not
% include URLs that reveal your institution or identity in your
% submission for review. Instead, provide an anonymous URL or upload
% the material as ``Supplementary Material'' into the CMT reviewing
% system. Note that reviewers are not required to look at this material
% when writing their review.

% \section*{Acknowledgements}

% \textbf{Do not} include acknowledgements in the initial version of
% the paper submitted for blind review.

% If a paper is accepted, the final camera-ready version can (and
% probably should) include acknowledgements. In this case, please
% place such acknowledgements in an unnumbered section at the
% end of the paper. Typically, this will include thanks to reviewers
% who gave useful comments, to colleagues who contributed to the ideas,
% and to funding agencies and corporate sponsors that provided financial

% \usepackage[backend=biber]{biblatex}

% \setcounter{biburllcpenalty}{7000}
% \setcounter{biburlucpenalty}{8000}

\bibliography{main}
\bibliographystyle{icml2019}


\clearpage
\appendix
\onecolumn

\setcounter{equation}{0}
\renewcommand{\theequation}{\thesection.\arabic{equation}}

\section{Proof of Proposition~\ref{prop:partial-order}}

From the definition of $C_\preceq(y)$,
$x \in C_\preceq(y)$ iff $x \preceq y$.
Then, we will show that
$x \in C_\preceq(y) \Leftrightarrow C_\preceq(x) \subseteq C_\preceq(y)$.

$(\Leftarrow)$ This is obvious because $x\in C_\preceq(x)$ holds.

$(\Rightarrow)$ For arbitrary $z\in C_\preceq(x)$,
$z \preceq x$ follows the definition of $C_\preceq(x)$.
Likewise, $x \preceq y$ follows $x \in C_\preceq(y)$.
Then, $z \preceq y$ holds because of the transitivity,
which implies that $C_\preceq(x) \subseteq C_\preceq(y)$.
$\square$

\section{Proof of Proposition~\ref{prop:convex-metric}}

\subsection{Non-negativity}
We will demonstrate this proposition by contradiction. Assume $d_W(\x, \y) < 0$; then, $s_j := \bm w_j^\top (\x - \y) < 0$ holds for all $j=1,\cdots,m$. From the assumption $\mathrm{coni}(W) = \Real^n$, there exists $a_1, \cdots, a_m \geq 0$ such that $\x - \y = \sum_{j=1}^m a_j \bm w$. Therefore,
\begin{align}
\label{eq:proof-l2-distance}
\| \x-\y \|^2 &= (\x-\y)^\top (\x-\y) \nonumber \\
&= \sum_{j=1}^m a_j \bm w_j^\top  (\x - \y) = \sum_{j=1}^m a_j s_j.
\end{align}
Considering $a_j\geq 0 $ and $s_j < 0$,
$\| \x-\y \|^2 < 0$ leads to a contradiction.
    
\subsection{Identity of indiscernibles}
If $d_W(x, y) = 0$, $s_j \leq 0$ holds for all $j=1,\cdots,m$.
Considering $a_j\geq 0 $ and $s_j \leq 0$ in \eqref{eq:proof-l2-distance},
we obtain $\| \x-\y \|^2 \leq 0$; then, $\x = \y$.

\subsection{Subadditivity}
\begin{align*}
d_W(\x, \y) &= \max_j \{ \bm w_j^\top (\x - \y) \} \\
&= \max_j \{ \bm w_j^\top (\bm x - \bm z) +   \bm w_j^\top (\bm z - \bm y) \} \\
&\leq \max_j \{ \bm w_j^\top (\bm x - \bm z) \}
    + \max_j \{ \bm w_j^\top (\bm z - \bm y) \} \\
&= d_W(\bm x, \bm z) + d_W(\bm z, \bm y). \;\; \square
\end{align*}

\section{Proof of Theorem~\ref{theo:equiv-ord-emb}}
Condition \eqref{eq:order-emb} is equivalent to
$\max_k \{x_k - y_k\} \leq 0$.
Thus, we will show that
$\max_k \{x_k - y_k\} = d(\bm x', \bm y') - r_x + r_y$
if
$\phi_\mathrm{ord}(\bm x) = (\bm x', r_x) , \, \phi_\mathrm{ord}(\bm y) = (\bm y', r_y)$.

Let $P_\perp = I - P = \frac 1n \bm 1 \bm 1^\top$; then,
\begin{align*}
\max_k \{x_k - y_k\}
&= \max_k \{\bm e_k^\top (\x - \y)\} \\
&= \max_k \{\bm e_k^\top (P+P_\perp) (\x - \y)\}.
\end{align*}
Here, considering
$P_\perp \bm e_k = \frac 1n \bm 1, \;
P \bm e_k = \bm w_k, \; P^2 = P$, we find
\begin{align}
\max_k \{x_k - y_k\}
&= \max_k \{ \bm w_k P(\x - \y)\}
+ \frac{\bm 1^\top \bm x}{n} - \frac{\bm 1^\top \bm y}{n}  \nonumber\\
&= \max_k \{ \bm w_k (\bm x' - \bm y')\} + (a - r_x) - (a - r_y)  \nonumber\\
&= d_W(\bm x', \bm y') - r_x + r_y. \;\; \square
\label{eq:proof-ord-equiv}
\end{align}

\section{Proof of Theorem~\ref{theo:loss-ord-emb}}

By using a uniform norm in \eqref{eq:order-loss}
instead of a Euclidean norm,
\begin{align}
\| h_+(\x - \y) \|_\infty
    &= \max_k \left\{\left| h_+(x_k - y_k) \right|\right\} \nonumber \\
    &= h_+\left(\max_k\left\{ x_k - y_k \right\}\right) \nonumber \\
    &= h_+\left(d_W \left(\bm x', \bm y'\right) - r_x + r_y \right) \nonumber \\
    &= h_+\left(l_{xy} \right),
    \label{eq:proof-theo2}
\end{align}
where $l_{xy} = l\left(\bm x', r_x; \bm y', r_y \right)$
and $h_+$ is applied element-wise.
We used \eqref{eq:proof-ord-equiv} for the third equation of \eqref{eq:proof-theo2}.

From the inequality between the uniform norm and the Euclidean norm
$\| \x \| \geq \| \x \|_\infty$, we find
\begin{equation*}
E^\mathrm{ord}(\x, \y) 
= \| h_+(\x - \y) \|^2
\geq  \| h_+(\x - \y) \|_\infty^2
= h_+ \left( l_{xy} \right)^2.
\end{equation*}
The equality holds iff
\begin{equation*}
    \| h_+(\x - \y) \| = \| h_+(\x - \y) \|_\infty,
\end{equation*}
i.e.,
\begin{equation*}
   \left|\left\{ k \middle| h_+(x_k - y_k) \neq 0 \right\}\right| \leq 1. 
   \; \; \square
\end{equation*}

%\section{Supplemental materials}
%
%In this document, we proof equivalences
%between our Disk Embeddings and existing methods;
%Order Embeddings \cite{Vendrov2016OEI}
%and Hyperbolic Entailment Cones \cite{Ganea2018HEC}.

\section{Proof of Theorem~\ref{theo:equiv-hyp-cones}}

We first prove Theorem~\ref{theo:loss-hyp-cones} and
then use our results to prove Theorem~\ref{theo:equiv-hyp-cones}.
Thus, see Sec.~\ref{subsec:proof:loss-hyp-cones} first.

By eliminating $d_x$ from \eqref{eq:x-norm} and \eqref{eq:coth-d-x},
we obtain
\begin{equation*}
\sin(r_x+\theta_0) =  \frac{1+\|\x\|^2}{2\|\x\|}\sin \theta_0,
\end{equation*}
which is followed by \eqref{eq:hyp-cones-map}.

The equivalence of ordering \eqref{eq:formal-disk} and \eqref{eq:hyp-cones} is
directly derived from Theorem~\ref{theo:loss-hyp-cones} since
\begin{equation}
E_{ij}^\mathrm{hyp} \leq 0 \Leftrightarrow l_{ij} \leq 0. \;\; \square
\end{equation}

\section{Proof of Theorem~\ref{theo:loss-hyp-cones}}
\label{subsec:proof:loss-hyp-cones}

%We show the derivation of equivalence
%between Disk Embeddings on $(n-1)$-sphere
%and Hyperbolic Entailment Cones on $n$-dimensional hyperbolic spaces.
%This can be considered as a proof of Theorem~\ref{theo:equiv-hyp-cones}
%and Theorem~\ref{theo:loss-hyp-cones}.

\begin{figure}[h]
    \vskip 0.2in
    \begin{center}
    \centerline{
        \includegraphics[width=3.5in]{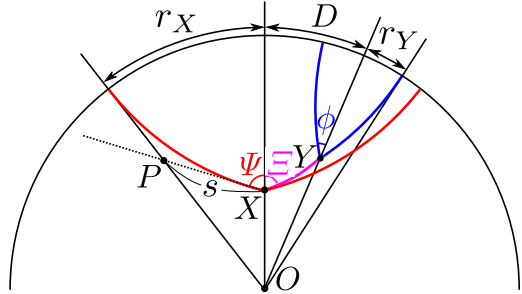}
    }
    \caption{
    Hyperbolic Entailment cones.
    \label{fig:hyp-cones}
    }
    \end{center}
    \vskip -0.2in
\end{figure}

To obtain \eqref{eq:loss-hyp-cones-equiv},
we present $\psi - \Xi$ in Figure~\ref{fig:hyp-cones} as a function of $r_X$, $r_Y$, and $D$.
Let $d_x$, $d_y$, and $d_{xy}$ be
\begin{eqnarray*}
    d_x & = & d_{\mathbb{D}}(O,X) \nonumber \\
    d_y & = & d_{\mathbb{D}}(O,Y) \nonumber \\
    d_{xy} & = & d_{\mathbb{D}}(X,Y) \nonumber
\end{eqnarray*}
and $x$ and $y$ be
\begin{align}
    x &= \| {\bm x} \| = \tanh \frac{d_x}{2}. \label{eq:x-norm} \\
    y &= \| {\bm y} \| = \tanh \frac{d_y}{2}. \label{eq:y-norm}
\end{align}

Assume that $XP = s$ in the Euclidean triangle $\triangle OPX$; then, $OP = 1 - s$. Thus,
\begin{equation}
    (1-s) \sin r_X = s \sin \psi. \label{eq:s-sin-psi}
\end{equation}

By applying the law of cosines to $\angle POX$, it is shown that
\begin{equation}
    s^2 = x^2 + (1-s)^2  - 2 x(1-s) \cos r_X. \label{eq:s-square}
\end{equation}

By removing $s$ from (\ref{eq:s-sin-psi}) and (\ref{eq:s-square}) and substituting (\ref{eq:x-norm}), we have  
\begin{equation}
    \sin \psi = \frac{\sin r_X}{\cosh d_x - \sinh d_x \cos r_X}. \label{eq:sin-psi}
\end{equation}

In addition, from the assumption of Hyperbolic Cones \cite{Ganea2018HEC},
\begin{equation}
    \sin \psi = K \frac{1 - x^2}{x} = \frac{2K}{\sinh d_x}. \label{eq:sin-psi-2}
\end{equation}

Comparing the right-hand side of equations (\ref{eq:sin-psi}) and (\ref{eq:sin-psi-2}), we have 
\begin{equation}
    \coth d_x = \cos r_X + \frac{1}{2K} \sin r_X
    = \frac{ \sin(r_X + \theta_0 )}{\sin \theta_0} \label{eq:coth-d-x}
\end{equation}
where $\theta_0 = \arctan 2K$.

In the same manner, we have
\begin{equation}
    \coth d_y = \frac{ \sin(r_Y + \theta_0 )}{\sin \theta_0}. \label{eq:coth-d-y}
\end{equation}

By substituting (\ref{eq:coth-d-x}) into (\ref{eq:sin-psi-2}),
\begin{align}
    \sin \psi &= 2K \sqrt{\coth^2 d_x -1} \nonumber\\
    &= \tan\theta_0 \sqrt{\frac{\sin^2(r_X-\theta_0)}{\sin^2\theta_0} - 1} \nonumber\\
    &= \frac{ \sqrt{\sin(r_X)\sin(r_X + 2\phi_0)}}{\cos \theta_0}. \label{eq:sin-psi-3}
\end{align}

Applying the law of sines and the law of cosines to the hyperbolic triangle $\triangle OXY$, we have 
\begin{align}
    \sinh d_y \sin D &= \sinh d_{xy} \sin \Xi \label{eq:sinh-d-y}, \\
    \cos D &= \frac{\cosh d_x \cosh d_y - \cosh d_{xy}}{\sinh d_x \sinh d_y}. \label{eq:cos-D}
\end{align}

By eliminating $d_{xy}$ from (\ref{eq:sinh-d-y}) and (\ref{eq:cos-D}),
and substituting (\ref{eq:coth-d-x}) and (\ref{eq:coth-d-y}), it is finally shown that 
%\begin{equation}
%    \psi - \Xi = \arcsin (\Xi_{ij} - \psi_i)
%\end{equation}
%where
\begin{align}
    \sin (\Xi -& \psi) =
2 \sin \!\left( \frac{r_X - r_Y - D}{2}\right) \,
\frac{
        \cos \left(\frac{r_X + r_Y - D}{2} + \theta_0 \right)
    }{
        \cos \theta_0 \, \sin \theta_0
    }
\sqrt{
    \frac{
        \sin r_X \sin(r_X+2 \theta_0)
    }{
        s_X^2 + s_Y^2 - 2 s_X s_Y \cos D - \sin^2 D
    }
}, 
\end{align}
%where $s_X = \sin(r_X+\theta_0), s_Y = \sin(r_Y+\theta_0)$.
where
$$s_X = \frac{\sin(r_X+\theta_0)}{\sin \theta_0},
s_Y = \frac{\sin(r_Y+\theta_0)}{\sin \theta_0}.$$

\section{Euclidean Entailment Cones}

Similar to Hyperbolic Cones \cite{Ganea2018HEC},
Euclidean Cones are also considered as Disk Embeddings.
Here, we will show that
$\psi-\Xi$ in Euclidean entailment cones is also represented by Rx, Ry and D.  

Let $d_x$, $d_y$, and $d_{xy}$ be  
\begin{eqnarray*}
	d_x = d(O,X) = x, \nonumber \\
	d_y = d(O,Y) = y, \nonumber \\
	d_{xy} = d(x,y). \nonumber
\end{eqnarray*}
(\ref{eq:euc-sin-psi-rx}), (\ref{eq:euc-sin-psi-ry}), and (\ref{eq:euc-sind}) are determined by applying the law of sines to $\triangle OAB$ and $\triangle OXY$:
\begin{align}
	\frac{\sin(\psi-R_x)}{x}=\sin\psi, \label{eq:euc-sin-psi-rx}\\
	\frac{\sin(\phi-R_y)}{y}=\sin\phi, \label{eq:euc-sin-psi-ry}\\
	\frac{\sin\Xi}{y}=\frac{\sin{D}}{d_{xy}}. \label{eq:euc-sind}
\end{align}\

Moreover, for Euclid entailment cones,
\begin{align}
    \sin\psi=\frac{K}{x} \label{eq:euc-sin-psi-k}, \\
    \sin\phi=\frac{K}{y} \label{eq:euc-sin-phi-k}. \\
\end{align}

By applying the law of cosines to $\triangle OXY$, we obtain $d_{xy}$:
\begin{eqnarray}
    {d_{xy}}^2 = x^2+y^2-2xy\cos{D}. \label{eq:euc-dxy}
\end{eqnarray}

We represent $\psi - \Xi$ as $r_X$, $r_Y$, $d_{xy}$, and $K$.
By eliminating $x$, $y$, $d_{xy}$, and $\phi$ from (\ref{eq:euc-sin-psi-rx}) to (\ref{eq:euc-dxy}), it is finally shown that
\begin{align}
    \sin(\psi &- \Xi) =
    	\frac{
        2 \sigma_X
        \cos\left(\frac{r_X - r_Y - D}{2}\right) 
        \sin\left(\frac{r_X + r_Y - D}{2} + \xi_0 \right)
    	}{\sqrt{
        \sigma_X^2 + \sigma_Y^2
        - 2 \sigma_X \sigma_Y \cos D
    }},
\end{align}
where $\sigma_X = \sin (r_X + \xi_0), \sigma_Y = \sin (r_Y + \xi_0)$
and $\xi_0 = \arcsin{K}$.

%\section{Marginal ReLU}
%\begin{figure}[t]
%    \vskip 0.2in
%    \begin{center}
%    \centerline{
%        \includegraphics[width=3.5in]{figures/relu}
%    }
%    \caption{
%    Marginal ReLU loss function.
%    \label{fig:relu}
%    }
%    \end{center}
%    \vskip -0.2in
%\end{figure}

%In Figure~\ref{fig:relu}, we illustrate the loss function
%we used and shown in Eq.\eqref{eq:loss-relu}.

\section{Loss functions for Hyperbolic Entailment Cones in Disk Embedding format}
\begin{figure}[h]
    \vskip 0.2in
    \begin{center}
    \centerline{
        \includegraphics[width=3.5in]{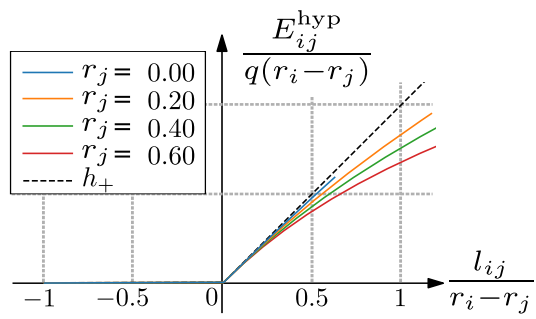}
    }
    \caption{
    Values of $E_{ij}^\mathrm{hyp}$ for $l_{ij}$ with fixed $r_i,r_j$.
    \label{fig:hyp-cones-loss}
    }
    \end{center}
    \vskip -0.2in
\end{figure}

In Figure~\ref{fig:hyp-cones-loss}, we illustrate 
values of energy function \eqref{eq:loss-hyp-cones-equiv}
for $l_{ij}$ with fixed $r_i,r_j$.

% We recommend that you build supplementary material in a separate document.
% If you must create one PDF and cut it up, please be careful to use a tool that
% doesn't alter the margins, and that doesn't aggressively rewrite the PDF file.
% pdftk usually works fine. 

% \textbf{Please do not use Apple's preview to cut off supplementary material.} In
% previous years it has altered margins, and created headaches at the camera-ready
% stage. 
%%%%%%%%%%%%%%%%%%%%%%%%%%%%%%%%%%%%%%%%%%%%%%%%%%%%%%%%%%%%%%%%%%%%%%%%%%%%%%%
%%%%%%%%%%%%%%%%%%%%%%%%%%%%%%%%%%%%%%%%%%%%%%%%%%%%%%%%%%%%%%%%%%%%%%%%%%%%%%%

\end{document}